\documentclass[conference]{IEEEtran}
\IEEEoverridecommandlockouts

\usepackage{cite}
\usepackage[english]{babel}
\usepackage[caption=false]{subfig}
\usepackage{amsmath,amssymb,amsfonts}
\usepackage{graphicx}
\usepackage{float}
\usepackage{booktabs}
\usepackage{siunitx}
\usepackage{multirow}
\usepackage[hidelinks]{hyperref}

\title{Offline Reinforcement Learning for Warehouse SLAM Throughput Control}

\author{
\IEEEauthorblockN{
Tina Dongxu Li,
Mouhacine Benosman,
Rajat Kumar,
Kevin Tan,
Ken Meszaros,
Trevor Dardik
}
\IEEEauthorblockA{
Amazon.com, USA \\
Emails: dxl@amazon.com, mbenos@amazon.com, kmardpl@amazon.com,\\
tkevita@amazon.com, mesza@amazon.com, tdardik@amazon.com
}
}

\begin{document}

\maketitle

\begin{abstract}
We present an offline reinforcement learning (RL) framework for optimizing SLAM throughput control in a warehouse fulfillment environment. SLAM (Scan/Label/Apply/Manifest) throughput directly influences system congestion and operational efficiency. Our RL-based control approach dynamically recommends SLAM throughput settings that adaptively balance throughput maximization with downstream stability through intelligent adjustment of throttling behavior. We include a history-informed state representation, action space abstraction for delayed-impact control, and a reward function that captures both upstream and downstream operational metrics. Our approach is algorithm-agnostic, enabling integration of multiple offline RL methods under a unified architecture. We instantiate our framework with three state-of-the-art offline RL algorithms, and trained the models offline using de-identified historical operational logs from a large-scale warehouse. Policy performance is evaluated using a comprehensive multi-method strategy. These include model-free approaches including immediate reward estimation via regression models and long-horizon Fitted Q Evaluation (FQE), as well as model-based Deep Koopman dynamics evaluation. Empirical results reveal that the CQL policy consistently outperforms alternatives, improving system health by $22.97\%$ and reducing average throttling duration by $3.18\%$. These findings demonstrate the potential of offline RL for safe and scalable warehouse throughput control optimization.
\end{abstract}

\begin{IEEEkeywords}
Offline Reinforcement Learning; Intelligent Control; Industrial Automation; Warehouse Automation; SLAM Throughput Optimization; Process Control
\end{IEEEkeywords}

\section{Introduction}

In large-scale warehouse fulfillment environments, SLAM systems are critical for labeling packages before directing them to downstream conveyor networks. When upstream processing rates exceed downstream handling capacity, congestion can occur, triggering protective control mechanisms that slow or pause conveyor movement. These congestion events negatively impact overall system efficiency and package processing times.

To mitigate congestion, upstream throughput can be regulated through throttling settings on automated SLAM lanes. However, excessive throttling may delay package processing and disrupt upstream workflows. This creates a challenging multi-objective optimization problem that requires balancing between upstream processing throughput and downstream congestion levels.

We employ an algorithm-agnostic Deep Reinforcement Learning (DRL) approach to learn adaptive throughput control policies for this problem. In doing so, we take an offline learning approach. This allows policies to be developed and evaluated safely using previously collected data, without direct interaction with the live control environment.
Our contributions include:
\begin{enumerate}

    \item \textbf{Temporal-aware state representation with predictive context: } 
The state design incorporates two temporal aspects: (a) multi-step historical signals of system throughput and congestion indicators to capture system dynamics, and (b) time-aligned upstream activity signals structured according to typical transit delays between processes. By incorporating these historical signals, the model develops an implicit understanding of package flow dynamics and delayed effects across upstream and downstream stages, and effectively anticipates future SLAM arrival rates, while enhancing adaptability without requiring separate forecasting models.

    \item \textbf{Transformation of the action space: }
Multi-level discrete throughput control configurations across multiple processing machines yield a prohibitively large action space. We aggregate these settings into time-weighted averages over fixed decision intervals, converting them into a compact continuous action space. This transformation exponentially reduces the dimensionality of the action space and better captures delayed system responses.

    \item \textbf{Balanced reward formulation:  }
We employ a custom-engineered reward function, in order to balance between upstream backlog indicators and downstream congestion signals. To improve learning stability, we normalize and rescale reward components to reduce scale disparities and improve gradient behavior during policy training.

    \item \textbf{Systematic cross-algorithm assessment architecture: }
We instantiate our approach with three popular offline RL algorithms (BCQ, CQL, and TD3+BC), each trained with multiple random seeds. Policy performance is assessed using a multi-metric evaluation strategy that includes immediate reward estimation via regression modeling, long-horizon value estimation using Fitted Q Evaluation (FQE), and action efficiency analysis based on throttling duration. In addition, we incorporate a model-based evaluation using a learned Deep Koopman dynamical system to simulate long-horizon policy rollouts under learned system dynamics. This combination of model-free and model-based evaluation methods enables a holistic assessment of policy performance and their applicability to warehouse process optimization.

\end{enumerate} 

Experimental results demonstrate the effectiveness of our proposed RL framework in addressing complex warehouse process optimization challenges. The CQL policy achieves the strongest performance among all evaluated policy alternatives, significantly improving the estimated long-term reward over the behavior policy while simultaneously reducing the average duration of throttling interventions. These results demonstrate the potential of offline reinforcement learning methods for data-driven warehouse process optimization.

\section{Related works}

Deep Reinforcement Learning (DRL) has been increasingly explored for optimizing warehouse operations. Most existing literature in this area studies the problems of single-robot navigation or multi-agent coordination in dynamic environments. For example, hierarchical multi-agent reinforcement learning (MARL) has been applied to coordinate mobile robots and human pickers in warehouse order-picking tasks, with a focus on spatial cooperation among heterogeneous agents ~\cite{krnjaic2024scalable}. Other research has explored the use of RL to improve item placement and retrieval efficiency for warehouse operations, by formulating it as a spatial Markov Decision Process ~\cite{cestero2022storehouse}. Although these studies apply RL to warehouse operational management, they primarily tackle spatial planning problems such as inventory placement and pick coordination. Our research addresses a fundamentally different challenge in warehouse operations: continuous automation control of conveyor systems.

Building upon these foundational studies, a small but growing body of work has begun to investigate applying DRL to process-level operational control. For example, researchers explored using reinforcement and transfer learning for dispatching decisions in manufacturing operations and showed how the RL based approach can dynamically release jobs to balance tardiness and inventory levels ~\cite{zheng2019manufacturing}. Another study proposed to use DRL to optimize the batching and sequencing of warehouse orders, with the aim of minimizing late deliveries and addressing when orders should be picked individually or assembled ~\cite{cals2020solving}. However, these existing studies focus on dispatching, sequencing, or picking decisions, without addressing the challenges of dynamic throughput control in warehouse management. 

Our work fills an important gap by introducing a deep reinforcement learning framework for SLAM throughput control in high-volume warehouse environments, specifically designed to handle delayed system-level rewards and feedback. 

While traditional DRL approaches are based on online learning, recent research demonstrates that offline reinforcement learning is a promising alternative that effectively learns policies without conventional online interactions and can be used to train robots, in both simulated and real environments ~\cite{kumar2019stabilizing, wu2019behavior, agarwal2020optimistic, kidambi2020morel, siegel2020keep}. Offline RL algorithms have the advantages of extracting optimal policies from massive logged datasets ~\cite{levine2020offline}. Traditional off-policy algorithms such as DQN and DDPG struggle with learning from fixed datasets due to extrapolation issues. To solve this problem, Batch-Constrained deep Q-learning (BCQ) uses a variational auto-encoder architecture to generate ``safe'' actions that are similar to those taken within the observed data \cite{fujimoto2019off}. Another approach, Conservative Q-Learning (CQL), prevents overestimation of Q-values by implementing a regularization mechanism that decreases the values of out-of-distribution actions and increases the values of observed actions ~\cite{kumar2020conservative}. Later, TD3+BC is introduced as a simpler approach that combines behavioral cloning and the TD3 algorithm ~\cite{fujimoto2021minimalist}. In our research, we use all three offline RL algorithms -- BCQ, CQL and TD3+BC -- within our RL approach.

In complex environments, state representations play a critical role in ensuring successful reinforcement learning. Existing work has discussed the augmentation of the state space with previous frames in the context of training deep reinforcement learning agents to play Atari games ~\cite{mnih2013playing}. In our warehouse SLAM throughput problem, we face a similar challenge of delayed system responses and delayed feedback. In our RL framework, we incorporate time-series data from upstream processes (e.g., pick and pack), capturing complex package flow dynamics and inter-process dependencies in a way not explored in previous literature on warehouse process optimization.

\section{Method}
\subsection{Problem formulation and objective}

To optimize warehouse fulfillment operations, an automated throughput control mechanism can be deployed to dynamically adjust the throttling settings of automated SLAM lanes in response to system conditions. We use ``SLAM throughput control'' and ``SLAM throttling'' interchangeably throughout this paper, where throttling refers to the operational mechanism used to regulate SLAM throughput. Previously, rule-based throughput control strategies have been implemented. However, such reactive heuristic approaches lack adaptability to evolving operational conditions and are often limited by human expertise, resulting in suboptimal decisions that prioritize short-term responses over long-term system stability. Despite these efforts, congestion still occurs frequently during high-volume periods. To address this persistent challenge, we propose a deep reinforcement learning framework to dynamically optimize SLAM throughput control. This data-driven approach learns policies that account for system state changes and optimize long-term system health rather than solely immediate outcomes.

We model this SLAM throughput control problem as a Markov Decision Process (MDP) defined by the tuple $(\mathcal{S}, \mathcal{A}, P, R, \gamma)$, where $\mathcal{S}$ is the state space that captures current system conditions, $\mathcal{A}$ is the action space that contains throttling configurations, $P(s'|s,a)$ denotes the system transition dynamics, $R(s,a)$ defines the reward function, and $\gamma \in (0,1)$ is the discount factor. At each decision interval $t$, the policy $\pi(a_t|s_t)$ selects an action $a_t \in \mathcal{A}$ based on the observed state $s_t \in \mathcal{S}$. The objective of the policy is to maximize the expected cumulative discounted reward \cite{Sutton2018}:
\begin{equation}
J(\pi) = \mathbb{E}_\pi \left[\sum_{t=0}^{\infty} \gamma^t r_t \right].
\end{equation}
In our SLAM throttling scenario, we aggregate state, action, and reward variables over fixed decision intervals rather than relying on instantaneous observations to filter out transient noise and ensure stable policy learning. Specifically, the state $s_t$ incorporates historical data from previous intervals, actions $a_t$ correspond to throttling settings in the future decision window, and the reward $r_t$ is collected during the subsequent evaluation window, reflecting the delayed impact of throttling decisions on the overall system performance. This comprehensive, temporally-aware architecture enables the reinforcement learning agent to capture the complex, dynamic relationships between control actions and their impacts on upstream and downstream system health.

\subsection{State representation}

The state in a MDP is not limited to the current observations, but can also include any knowledge the agent has about the surrounding environment. Future processing volume at downstream SLAM stations is largely influenced by upstream workflow activities. Downstream throttling decisions also indirectly influence upstream throughput when excessive throttling slows the overall pipeline. The agent benefits from visibility into recent upstream activities in order to make informed decisions. Therefore, we augment state space with aggregated operational metrics over multiple previous time intervals for each major upstream process (e.g. pick, pack). The historical interval length of each process to be included in the state is selected based on an empirical analysis of lag relationships within the training dataset. 

By directly integrating these raw signals into the policy input, it allows the model to implicitly learn the temporal dependencies and latent structure that link upstream activities to downstream SLAM volume and reward outcomes, so that the policy is able to condition its decision on anticipated trends in package arrival rate, without the need of a separate explicit volume prediction model. 

\subsection{Action design}

Each SLAM machine has multiple discrete throughput control levels, each implemented via a corresponding throttling duration between package releases. At a decision timestep $t$, machine $i$ can choose to switch between $M$ discrete throttling levels or not. Each throttling level $k$ lasts for a duration $d_{i,k,t}$, and corresponds to a throttling value $a_k$.

We aggregate each machine’s throttling settings within fixed decision intervals using a time weighted average, in order to aggregate a high-dimensional discrete action space into a low-dimensional continuous action space. The aggregated throttling action for machine $i$ is a time-weighted average:
\begin{equation}
\tilde{a}_{i,t} = \frac{\sum_{k=1}^{M} d_{i,k,t} \, a_k} {\sum_{k=1}^{M} d_{i,k,t}}
\end{equation}

During model training, we use the aggregated value $\tilde{a}_{i,t}$ as the continuous action representation for each SLAM machine, which dramatically reduces the action space from a cardinality of $M^N$ to a dimension of $N$. During model inference, we convert the recommended continuous actions back to the original $M$ discrete throttling levels by matching each action to its closest throttling setting. 

This action transformation is employed for two reasons. Firstly, our model must recommend optimal throttle settings for $N$ SLAM machines at any given time, and each machine has $M$ different throttle levels. As a result, the action space grows exponentially (e.g. $M^N$). This is a completely intractable action space, and so we employ a smaller representation in order to avoid the curse of dimensionality.
Secondly, the throttling actions have a delayed impact/reward to the system, and the impact/reward is accumulated over time. If actions are modeled with changes that are too frequent or durations that are too short, the impact of delayed feedback worsens and may cause instability.

\subsection{Reward design}

Effective throttling should balance downstream congestion mitigation with maintaining sufficient upstream throughput. Excessive throttling can reduce downstream congestion but may also slow upstream processing activities and eventually negatively impact system throughput. Therefore, we propose a balanced reward metric that captures a trade-off between downstream congestion indicators and upstream backlog signals, with adjustable weights assigned to each reward component. 
Several techniques are applied to further improve the reward design:

\begin{enumerate}
    \item \textbf{Selection of rolling window length:} The length of this rolling window must to be carefully selected. If the window is too short, it may not capture the majority of upstream backlog counts and would also have large variations in backlog counts (noisy reward signal); but if the window is too long, the backlog counts would be close to a uniform distribution and the reward will not be sensitive enough to policy actions, so the agent will not be able to learn. The approach we use is to look at the statistical distribution of historical cycle times from upstream to downstream processes and use the 95th percentile value as the rolling window length for upstream backlog indicator calculation.

    \item \textbf{Scaling the upstream backlog count:} The downstream congestion indicator itself is recorded as a percentage value in dataset, and the upstream backlog count can be derived by subtracting the number of SLAM processed packages (SLAMed) by the number of Packed packages over a rolling window. However, these two reward components are at significantly different scale and would cause issues like unstable gradient updates and dominance of larger-scale component which would eventually lead to sub-optimal policies. Instead of directly using the upstream backlog count, we convert it into a percentage value using: 
$
\text{Upstream Backlog Rate} = 
\frac{\# \text{Packed} - \# \text{SLAMed}}{\# \text{Packed}}.
$


    \item \textbf{Blending the two reward components: } The weights assigned to each of the two reward components can be treated as hyper-parameters and tuned during model training. 
The final reward is defined as:
\begin{equation}
\textit{reward} = w_1* \textit{R\_upstream} + w_2 * \textit{R\_downstream},
\end{equation}
where $w_1$ is the weight for the upstream reward, reflected by the upstream backlog rate; $w_2$ is the weight for the downstream reward, represented by the downstream congestion rate; $w_1 + w_2 = 1$.
\end{enumerate}
The choice of discount factor $\gamma$ is closely related to practical considerations regarding the interaction among the reward function, agent, and environment. The discount factor should be appropriately set based on the typical time horizon over which control actions influence system outcomes and the decision interval length. For example, with a 15-minute decision interval and $\gamma = 0.6$, $\sum_{t=0}^{\infty} \gamma^t$ captures nearly 90\% of the cumulative effects of an action within the first hour. The value of the discount factor $\gamma$ also directly affects the length of the rollout trajectory used in our model-based simulation-driven evaluation, and therefore influences its variance.

\subsection{Algorithm-agnostic offline policy learning}

Our framework is algorithm-agnostic in the sense that it allows different offline RL algorithms (TD3+BC, CQL, and BCQ) to be integrated without customized implementations. Concretely, we expose a unified interface that takes as input a fixed dataset $\mathcal{D}=\{(s,a,r,s')\}$, optional constraints $\mathcal{C}$, and training utilities (early stopping, evaluation), and returns a policy $\pi_\theta$ and critic $Q_\psi$.

Regardless of the chosen algorithm, the critic is trained with a one-step temporal difference (TD) regression target\cite{mnih2013playing}:
\begin{equation}
\begin{aligned}
\mathcal{L}_Q(\psi)&=\mathbb{E}_{(s,a,r,s')\sim\mathcal{D}}\!\left[\big(Q_\psi(s,a)-y\big)^2\right],\\
y&=r+\gamma\,\mathbb{E}\big[\bar Q_{\bar\psi}(s',a')\big],\ a'\sim \pi_{\bar\theta}(s').
\end{aligned}
\end{equation}

All algorithms share the same experimental pipeline, including feature engineering and a consistent set of model-free and model-based evaluation methods to enable fair and reproducible comparison across different offline RL algorithms.

\section{Experiments}

\subsection{Experimental Environment and Test Scenario}
The experimental environment is constructed from de-identified historical event logs collected from a large-scale warehouse SLAM system. The test scenario evaluates adaptive SLAM throughput control under varying warehouse volume conditions, where throughput control actions balance upstream package flow and downstream congestion across multiple automated SLAM lanes. The dataset includes timestamped records of upstream processes (e.g., pick and pack), SLAM lane throughput configurations, and downstream congestion indicators. Each decision step corresponds to a fixed 15-minute control interval, during which throttling actions are assumed to remain constant. System states are constructed from aggregated statistics over multiple past intervals, aligned using event timestamps to capture delayed process propagation effects. Since no online interaction with the live warehouse system is performed, the environment is fully offline. Policy evaluation is conducted using held-out historical trajectories together with model-free and model-based offline evaluation methods. The model-based evaluation rolls out trained policies within learned system dynamics models rather than a live production environment. All experiments and evaluations were implemented in Python within an offline research environment on remote cloud computing infrastructure.

\subsection{Evaluation Overview}
We conducted comprehensive experiments to assess the validity and generalization ability of our proposed RL framework. Given that our state vector incorporates historical time steps and we prepare state-action-reward sequences using a sliding window approach, the samples naturally contain overlapping time intervals. Consequently, a random split of the data into training and testing sets would introduce information leakage. To avoid this issue, we performed a temporal split, assigning the earlier portion of the data to training and the later portion to testing. This resulted in 65,088 training samples and 21,510 test samples. We implemented our RL framework using three state-of-the-art offline RL algorithms: BCQ, CQL, and TD3+BC, on five different random seeds. 

To comprehensively evaluate policy performance, we adopt both model-free and model-based evaluation approaches. Model-free evaluation includes immediate reward estimation, Fitted Q Evaluation (FQE) for long-term discounted value estimation,  and throttling action efficiency analysis. Model-based evaluation is implemented with a Deep Koopman method to learn the system dynamics in order to assess long-horizon policy behavior. 

In addition, we conduct robustness analyses across different dataset sizes, volume modes, discount factors, and reward weighting strategies. Throttling duration values are linearly normalized to the range [0, 1] for visualization. Detailed descriptions of each evaluation method are provided in the following subsections.

\subsection{Model-free evaluation}

This section presents the model-free evaluation results, including immediate reward estimation, FQE, and action efficiency metrics. The results indicate that the CQL, BCQ, and TD3+BC policies all achieved positive FQE scores and reductions in throttling duration, indicating that these policies consistently achieved higher long-term rewards than the behavior baseline while simultaneously reducing average throttling duration, with findings remaining stable across all five seeds. Table~\ref{tab:model_free} summarizes the percentage improvement in performance metrics of each RL policy relative to the behavior baseline, averaged across five random seeds. CQL stands out as the top performer with a FQE improvement of 22.97\% and throttling reduction of 3.18\% on average. Evaluation of immediate rewards revealed both positive and negative results across all algorithms, indicating that RL policies do not always optimize immediate outcomes. This behavior aligns with the problem’s objective, which prioritizes long-term cumulative performance over immediate gains.

\begin{table}[H]
\centering
\caption{Model-Free Evaluation on Different RL Policies}
\label{tab:model_free}
\begin{tabular}{lccc}
\toprule
& FQE & Immediate Reward & Throttling Reduction \\
\midrule
CQL         & +22.97\% & -114.20\% & -3.18\%  \\
BCQ         & +10.49\% &   +0.64\%  & -7.65\%  \\
TD3+BC      &  +9.90\% &  -2.93\%  & -0.57\%  \\
\bottomrule
\end{tabular}
\end{table}

\subsection{Model-Based Evaluation with Deep Koopman}
To complement the model-free evaluation, we also conducted model-based policy evaluation using a learned simulation environment, where each policy was rolled out for 8 decision steps (corresponding to an extended control horizon) and average discounted return was computed. Here, the transition dynamics are heavily nonlinear -- a simple linear state transition model incurs substantial prediction error, with an 8-step Weighted Absolute Percentage Error (WAPE) of 52.31\%. To improve long-horizon simulation fidelity, we instead trained a DeepKoopmanWithAR model, a neural Koopman-based dynamical system that learns a linear latent representation of nonlinear dynamics with an autoregressive structure~\cite{lusch2018deep}. The DeepKoopmanWithAR environment reduced the WAPE to 22.25\%, representing a significant improvement over the linear baseline. Simulation results from this environment show that CQL achieved the highest return, substantially outperforming TD3+BC, Behavior Policy, and BCQ. Table~\ref{tab:model_based} reports normalized performance scores by linearly scaling each policy’s average discounted return with the formula:
\begin{equation}
\text{Normalized Score}_i =
\frac{R_i - R_{\min}}{R_{\max} - R_{\min}}
\end{equation}
where $R_i$ is the average discounted return of policy $i$, $R_{\max}$ and $R_{\min}$ are the maximum and minimum returns among all evaluated policies. Overall, both the model-based DeepKoopman evaluation and model-free FQE evaluation provide consistent evidence that CQL is the strongest candidate policy.

\begin{table}[H]
\centering
\caption{Model-Based Evaluation on Different Policies}
\label{tab:model_based}
\begin{tabular}{l r}
\toprule
\textbf{Method} & \textbf{Normalized Score} \\
\midrule
CQL             & 1.00 \\
BCQ             & 0.64 \\
Behavior Policy & 0.27 \\
TD3+BC          & 0.00 \\
\bottomrule
\end{tabular}
\end{table}

\subsection{Policy Performance Under Varying Training Data Sizes}

\begin{figure}[H]
    \centering
    \includegraphics[width=\columnwidth]{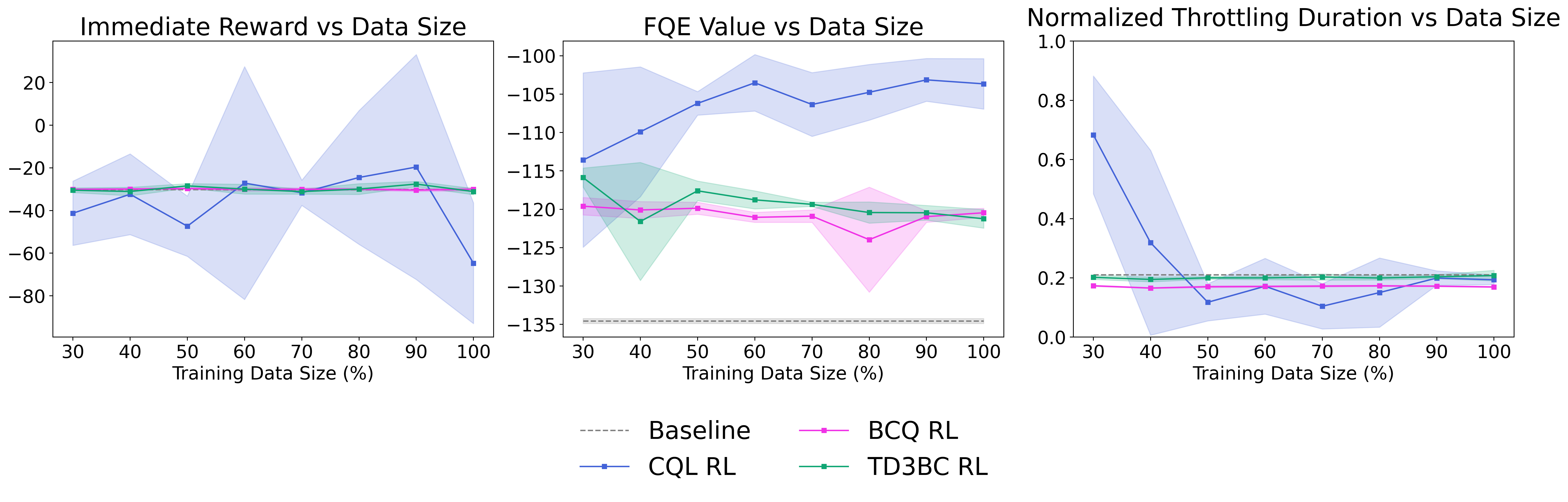}
    \caption{Evaluation on different training data sizes}
    \label{fig:cross_datasize}
\end{figure}

We evaluated the impact of data availability on policy performance by training with varying proportions of the training dataset and evaluating on a fixed test set. As shown in Fig.~\ref{fig:cross_datasize}, CQL exhibits clear performance gains as the training dataset size increases, reflected by higher FQE values and reduced average throttling duration. In contrast, TD3+BC and BCQ demonstrate relatively stable performance across different data sizes. Although such stability may reflect robustness, it also suggests a more limited capacity to leverage additional data coverage for further policy improvement. In many large-scale operational environments, historical [state, action, reward] logs can accumulate over time, enabling progressive dataset expansion. In such settings, algorithms that systematically benefit from increased data availability are particularly advantageous. Given its strong data scalability, along with superior performance in both model-free and model-based evaluations, CQL is selected as the primary candidate for subsequent analysis.

\subsection{Performance Across Volume Modes}

\begin{figure}[H]
    \centering
    \includegraphics[width=0.6\columnwidth]{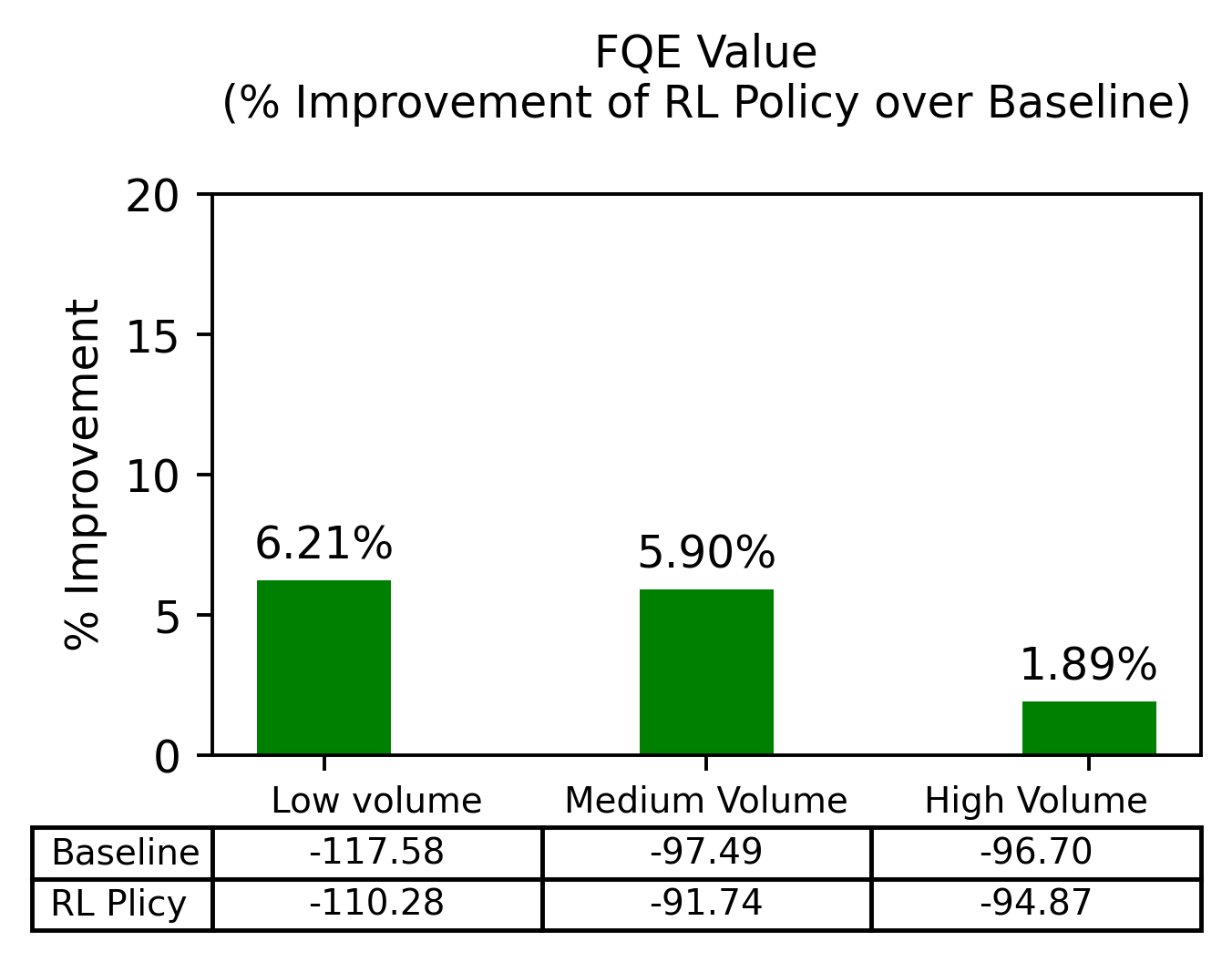}
    \caption{FQE score of CQL policy across volume modes}
    \label{fig:fqe_by_volume}
\end{figure}

\begin{figure}[H]
    \centering
    \includegraphics[width=\columnwidth]{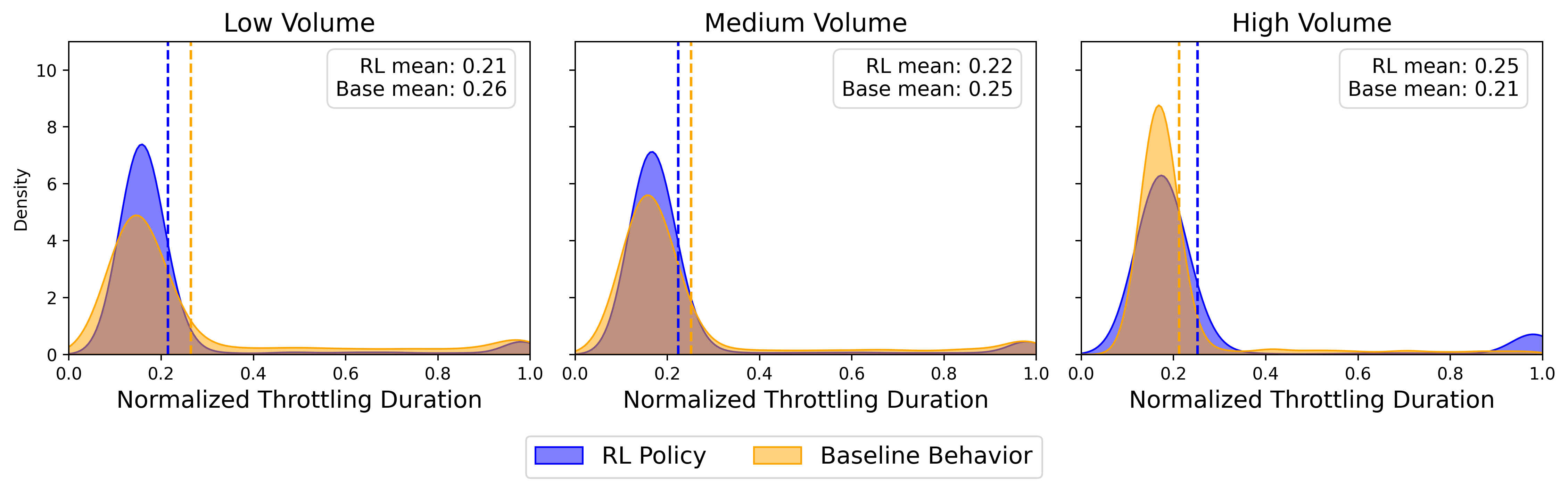}
    \caption{Action distribution of CQL policy across volume modes}
    \label{fig:throttle_by_volume}
\end{figure}

We conducted further analysis to examine how the CQL policy performs under varying operating conditions—low, medium, and high volume modes. CQL consistently improves FQE across all volume modes, achieving performance gains of 6.21\%, 5.90\% and 1.89\% respectively (Fig.~\ref{fig:fqe_by_volume}). Under low and medium volume modes, the policy recommends shorter throttling durations than the behavior baseline, suggesting the learned policy improves system efficiency while avoiding unnecessary throttling. In contrast, under high volume conditions, the policy adopts a more conservative throughput control strategy by recommending longer throttling durations to maintain system stability and prevent congestion (Fig.~\ref{fig:throttle_by_volume}). The results indicate CQL learns volume-sensitive control behavior, dynamically adjusting throttling in response to different throughput levels to optimize overall system performance.  

\subsection{Ablation Study}
\begin{figure}[H]
    \centering
    \includegraphics[width=\columnwidth]{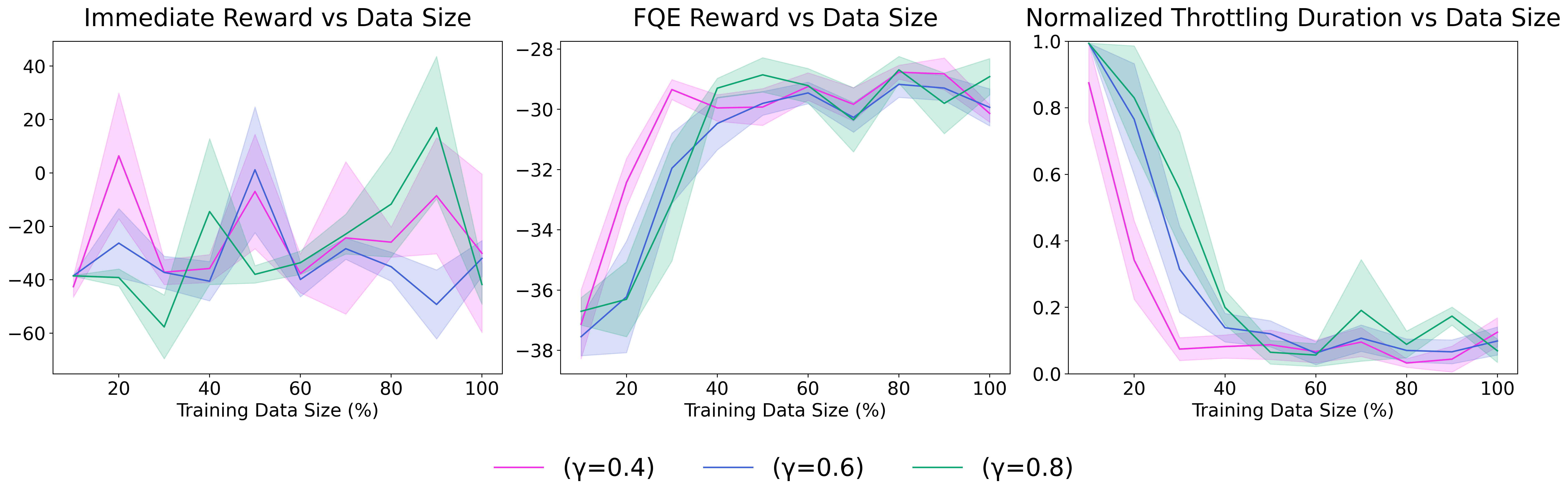}
    \caption{Evaluation on Different Discount Rates}
    \label{fig:cross_gamma}
\end{figure}

\begin{itemize}
    \item Discount Factor Sensitivity
\end{itemize}
The choice of reward discount factor ($\gamma$) affects the effective temporal planning horizon of the policy. For example, if each decision step represents a 15-min interval, when $\gamma$ = 0.4, approximately 84\% of the cumulative discounted impact is concentrated within the next 30 minutes; when $\gamma$ = 0.6, approximately 87\% is concentrated within the next 60 minutes; and when $\gamma$ = 0.8, approximately 84\% is concentrated within the next two hours. We trained CQL using different $\gamma$ values and plotted the results in Fig.~\ref{fig:cross_gamma}. To enable fair comparison across discount factors, all FQE values were rescaled by multiplying by $(1-\gamma)$, which normalizes the magnitude of discounted returns. Results show that CQL is robust to discount factor selection and maintains consistent performance across planning horizons.    

\begin{figure}[H]
    \centering
    \includegraphics[width=\columnwidth]{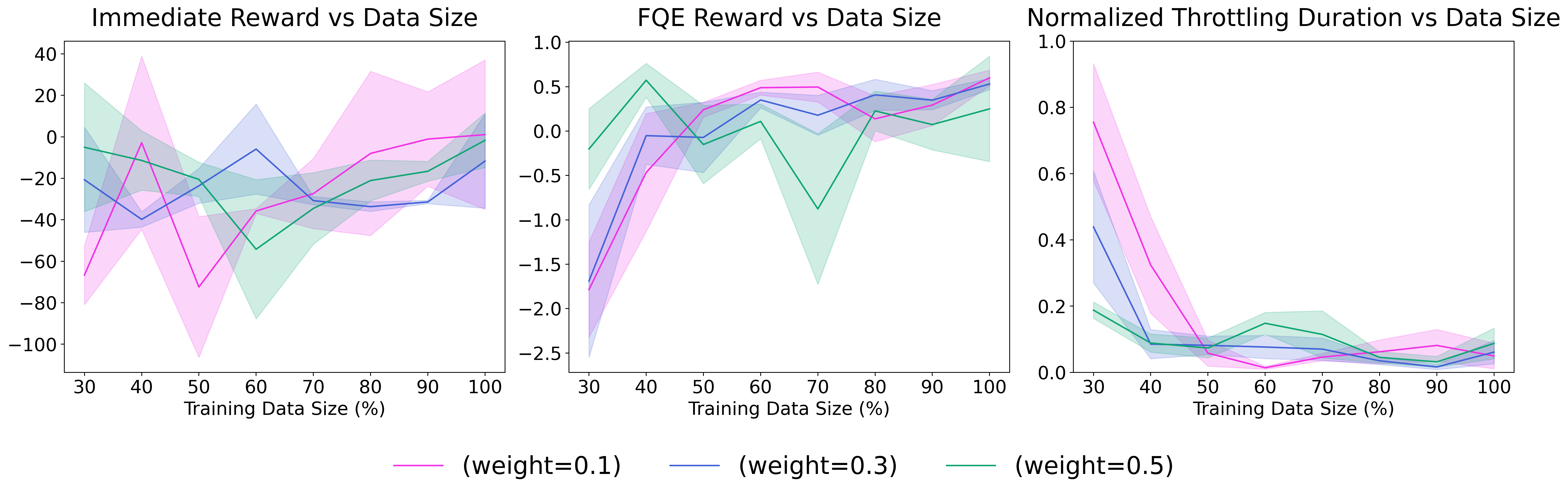}
    \caption{Evaluation on Different Reward Weighting Strategies}
    \label{fig:cross_weight}
\end{figure}

\begin{itemize}
    \item Reward Weighting Sensitivity
\end{itemize}
The reward metric is a weighted sum of upstream backlog indicators and downstream congestion indicators. The RL framework balances these two reward components through adjustable weights. Since excessive downstream accumulation can propagate upstream and impact overall system dynamics, controlling downstream fullness should generally be prioritized. We evaluated three weighting strategies and results are shown in Fig.~\ref{fig:cross_weight}. When the upstream weight was set to 0.1 or 0.3, both FQE and throttling performance improved consistently as dataset size increased, indicating stable learning. In contrast, higher upstream weighting (0.5 or above) resulted in less stable policies and weaker learning gains. This aligns with operational logic, as overemphasizing upstream WIP reduction can compromise downstream stability, which has a greater system-wide impact. Based on these experimental findings, assigning a higher weight to the downstream component yields the most stable policy behavior.

\section{Conclusion}
In this study, we developed a deep reinforcement learning framework to optimize SLAM throughput control in the warehouse environment, addressing the critical trade-off between maximizing throughput and maintaining downstream system stability. By leveraging a history-informed state representation, an action abstraction strategy to handle delayed impacts, and a balanced reward function that captures both upstream and downstream operational metrics, our deep RL framework learns effective control policies. These results were consistently validated across multiple random seeds and under different evaluation strategies, including model-free approaches such as immediate reward regression models, long-term FQE analysis and model-based DeepKoopman evaluation. The CQL algorithm achieves particularly substantial improvements over baseline operations. The model-free FQE evaluation demonstrates that the CQL policy improved long-term cumulative rewards by 22.97\%, while simultaneously reducing average throttling duration by about 3.18\%, indicating enhanced operational efficiency with less intervention required. These findings demonstrate the promise of leveraging a data-adaptive reinforcement learning framework to optimize complex warehouse processes such as SLAM throughput control. 

\appendix
\section{Appendix A}
\subsection{Comparative study of training convergence: BCQ, CQL, and TD3+BC}
\begin{figure}[H]
    \centering
    \includegraphics[width=\columnwidth]{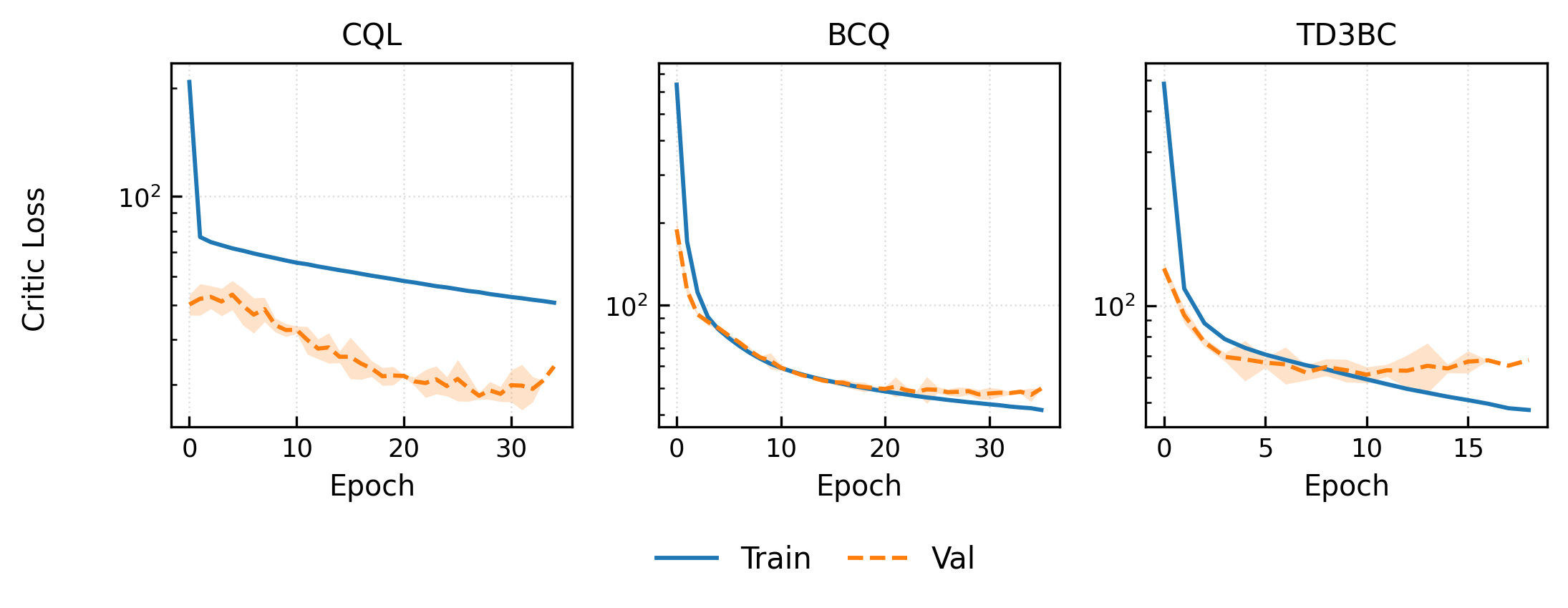}
    \caption{Training and validation critic loss curves averaged across five random seeds for the three offline RL algorithms. Shaded regions represent ±2 standard deviations.}
    \label{fig:loss_curve}
\end{figure}

\subsection{Evaluation tables}

\begin{table}[H]
\centering
\caption{Percentage improvement in average FQE value of each RL policy relative to the baseline across random seeds.}
\label{tab:fqe-improvement}
\small
\begin{tabular}{l 
  S[table-format=2.2, table-number-alignment=right]
  S[table-format=2.2, table-number-alignment=right]
  S[table-format=2.2, table-number-alignment=right]}
\toprule
& {CQL (\%)} & {BCQ (\%)} & {TD3+BC (\%)} \\
\midrule
seed0  & 22.72 & 10.43 &  8.94 \\
seed1  & 19.42 & 10.94 & 10.66 \\
seed2  & 21.66 & 10.59 &  8.56 \\
seed3  & 24.17 & 10.11 & 10.63 \\
seed4  & 26.88 & 10.37 & 10.72 \\
\midrule
Std    &  2.79 &  0.31 &  1.06 \\
\bfseries Mean 
       & \bfseries 22.97 & \bfseries 10.49 & \bfseries 9.90 \\
\bottomrule
\end{tabular}
\end{table}

\begin{table}[H]
\centering
\caption{Percentage change in average throttling duration of each RL policy relative to the baseline across random seeds.}
\label{tab:throttling-duration-change}
\small
\begin{tabular}{l 
  S[table-format=-1.2, table-number-alignment=right]
  S[table-format=-1.2, table-number-alignment=right]
  S[table-format=-1.2, table-number-alignment=right]}
\toprule
& {CQL (\%)} & {BCQ (\%)} & {TD3+BC (\%)} \\
\midrule
seed0  &  -1.71 &  -8.05 &  -3.94 \\
seed1  &  -1.77 &  -7.56 &   0.01 \\
seed2  &  -9.14 &  -7.83 &  -1.31 \\
seed3  &  -1.78 &  -7.30 &   5.38 \\
seed4  &  -1.49 &  -7.51 &  -2.98 \\
\midrule
Std    &   3.34 &   0.29 &   3.66 \\
\bfseries Mean 
       & \bfseries -3.18 & \bfseries -7.65 & \bfseries -0.57 \\
\bottomrule
\end{tabular}
\end{table}

\begin{table}[H]
\centering
\caption{Percentage change in average immediate reward of each RL policy relative to the baseline across random seeds.}
\label{tab:immediate-reward}
\small
\begin{tabular}{l 
  S[table-format=-3.2, table-number-alignment=right] 
  S[table-format=-2.2, table-number-alignment=right] 
  S[table-format=-2.2, table-number-alignment=right]}
\toprule
& {CQL (\%)} & {BCQ (\%)} & {TD3+BC (\%)} \\
\midrule
seed0  & -217.40 &   2.01 &  -2.36 \\
seed1  &  -71.61 &  -1.97 &   0.15 \\
seed2  &   17.01 &  -2.34 &   0.57 \\
seed3  &  -73.23 &   4.08 &  -0.58 \\
seed4  & -225.75 &   1.41 & -12.42 \\
\midrule
Std    &  104.65 &   2.74 &   5.42 \\
\bfseries Mean 
       & \bfseries -114.20 & \bfseries 0.64 & \bfseries -2.93 \\
\bottomrule
\end{tabular}
\end{table}

\section*{ACKNOWLEDGMENT}
The authors would like to thank the leadership team for their support and guidance throughout this project. During the preparation of this work, the authors used Amazon Q to improve spelling, grammar, and readability of the text.

\bibliography{references}
\bibliographystyle{IEEEtran}

\end{document}